\documentclass[conference,9pt]{IEEEtran}
\IEEEoverridecommandlockouts

\usepackage{cite}
\usepackage{amsmath,amssymb,amsfonts}
\usepackage{algorithmic}
\usepackage{graphicx}
\usepackage{textcomp}
\usepackage{xcolor}
\def\BibTeX{{\rm B\kern-.05em{\sc i\kern-.025em b}\kern-.08em
    T\kern-.1667em\lower.7ex\hbox{E}\kern-.125emX}}

\usepackage{bm}
\usepackage[hidelinks]{hyperref}
\usepackage{multirow}
\usepackage{makecell}
\usepackage{setspace}
\usepackage{booktabs}

\usepackage{tikz}
\usetikzlibrary{calc}
\usetikzlibrary{positioning}
\usetikzlibrary{shapes.geometric}
\usepackage{ifthen}

\usepackage{amssymb}\usepackage{pifont}\newcommand{\cmark}{\ding{51}}\newcommand{\xmark}{\ding{55}}

\usepackage{pgfplots}
    \pgfplotsset{compat=newest}
    \usetikzlibrary{plotmarks}
    \usepgfplotslibrary{groupplots}

\definecolor{mygreen}{RGB}{228,240,218}
\definecolor{myblue}{RGB}{220,228,241}

\def\x{{\mathbf x}}

\begin{document}

\title{Zero-resource Speech Translation and Recognition with LLMs\thanks{Accepted at ICASSP 2025.}}

\author{\IEEEauthorblockN{
Karel Mundnich, 
Xing Niu, 
Prashant Mathur, 
Srikanth Ronanki, 
Brady Houston,
Veera Raghavendra Elluru,\\
Nilaksh Das,
Zejiang Hou, 
Goeric Huybrechts, 
Anshu Bhatia,
Daniel Garcia-Romero, 
Kyu J. Han, 
Katrin Kirchhoff}
\IEEEauthorblockA{\textit{AWS AI Labs} \\
\{kmundnic, xingniu, ronanks\}@amazon.com}
}

\maketitle

\begin{abstract}
Despite recent advancements in speech processing, zero-resource speech translation (ST) and automatic speech recognition (ASR) remain challenging problems. In this work, we propose to leverage a multilingual Large Language Model (LLM) to perform ST and ASR in languages for which the model has never seen paired audio-text data. We achieve this by using a pre-trained multilingual speech encoder, a multilingual LLM, and a lightweight adaptation module that maps the audio representations to the token embedding space of the LLM. We perform several experiments both in ST and ASR to understand how to best train the model and what data has the most impact on performance in previously unseen languages. In ST, our best model is capable to achieve BLEU scores over 23 in CoVoST2 for two previously unseen languages, while in ASR, we achieve WERs of up to 28.2\%. We finally show that the performance of our system is bounded by the ability of the LLM to output text in the desired language.
\end{abstract}


\section{Introduction}
The goal of zero-resource speech processing is to learn from unlabeled audio (i.e., audio that is not paired to text), similar to what humans do in their infancy before acquiring the ability to read or write \cite{dunbar2022self}. Even with a good deal of efforts over the past decades, zero-resource ASR or ST are still challenging tasks in the more general case (learning from unpaired text and audio), leading to work that splits the general zero-resource problem into simpler tasks \cite{dunbar2022self}. In this work, we focus on the problem of cross-lingual transfer between \textit{seen} and \textit{unseen} languages in the audio data during training, to perform speech processing tasks in \textit{unseen} languages at test time by leveraging cross-lingual transfer and multilingual LLMs.

Recently, speech LMs (SLMs) have shown remarkable performance in speech processing tasks \cite{gao2022wavprompt, fathullah2023prompting, wang2023slm, chu2023qwen, team2023gemini} by leveraging both pre-trained audio encoders \cite{baevski2020wav2vec} as well as the tremendous progress and generalization capabilities of LLMs \cite{kojima2022large}. The success of these new models can be partly explained by the fact that pre-training techniques are enabling bridging the gap between the vast amounts of unlabeled audio and text. This has led to state-of-the-art performance in several tasks including multilingual ASR \cite{wang2023slm, team2023gemini} and speech translation \cite{tsiamas2024pushing}.

Despite the successes of SLMs, most of the work has focused on understanding the generalization capabilities of instruction-tuned LLMs in supervised speech processing tasks, where the spoken languages used at test time are seen during the training of the SLM. Therefore, little to no effort has been put into understanding zero-resource cross-lingual transfer to \textit{unseen} audio languages when the SLM is an end-to-end model composed of a multilingual pre-trained speech encoder and a fine-tuned LLM. For example, in \cite{rubenstein2023audiopalm} the authors study the zero-shot capabilities of the model in ST for unseen \textit{pairs} of languages at training time, but all the input languages were used at training time for at least one of the tasks (so the model has seen labeled data for all the input languages). Two other works focus on zero-shot translation from English to other languages, so that the \textit{pairs} of languages are previously unseen (but the input is fixed to English). In \cite{tsiamas2024pushing}, the authors propose a technique to fine-tune a pre-trained speech encoder using CTC and an optimal transport loss to map the speech representations directly into the text token embedding space, achieving state-of-the-art results in ST using a multilingual LLM, while in \cite{wang2022discrete}, the authors perform zero-shot translation from English to several other languages by proposing a model architecture to find a shared semantic space for audio and text representations. Finally, in \cite{klejch2021deciphering} the authors propose a method for zero-resource ASR using multilingual transfer, by pre-training an audio encoder and using a universal phone recognizer trained on a different set of languages than the test languages and a deciphering algorithm.

In this paper, we study the capacity of a multilingual LLM to perform zero-resource ST and ASR on languages for which we use no labeled audio data. To achieve this, we pre-train a speech encoder with unlabeled data from over 130 languages, and train an SLM with a lightweight adapter that maps the audio representations to the token embedding space of the LLM. We show that with this approach, it is possible to perform zero-resource ST from previously-unseen languages to English, and to perform ASR in these same unseen languages.

Our contributions are the following:
(1) we propose a new approach to perform zero-resource ST and ASR through cross-lingual transfer by leveraging multilingual LLMs;
(2) we study the impact of multitask training (using ST and ASR as tasks) and sequential training in the zero-resource performance;
(3) we show that the ability of the LLMs to generate text in the target language is crucial for the ASR performance.

\section{Methodology}
\label{sec:methodology}

\begin{figure}[t]
    \centering
    \clearpage{}    \begin{tikzpicture}[scale=0.7,
        transform shape,
llm/.style={
            rectangle,
            draw=black,
            thick,
            fill=mygreen,
            align=center,
            rounded corners,
            font=\small,
            minimum height = \columnwidth/8,
},
        lora/.style={
            rectangle,
            draw=black,
            thick,
            fill=yellow!30,
            rounded corners,
            font=\small,
            minimum width=2cm,
            minimum height=1cm,
            align=right,
        },
        ttoken/.style={
            rectangle,
            draw=black,
            thick,
            fill=mygreen,
            align=center,
            rounded corners,
            minimum height = 3,
            minimum width = 4,
        },
        atoken/.style={
            rectangle,
            draw=black,
            thick,
            fill=myblue,
            align=center,
            rounded corners,
            minimum height = 3,
            minimum width = 4,
        },
        textbox/.style={
            rectangle,
            draw=black,
            dotted,
            fill=white,
            align=center,
        },
        audiobox/.style={
            rectangle,
            draw=black,
            dashed,
            fill=white,
            align=center,
        },
        myarrow/.style={
            -stealth,
            draw=black,
            thick,
        },
        myline/.style={
            draw=black,
            thick,
        },
        elbowline/.style={
            draw=black,
            thick,
        },
        dot/.style={
            circle,
            fill=black,
            inner sep=1pt
        },
        cnn/.style={
            rectangle,
            draw=black,
            thick,
            fill=myblue,
            align=center,
            rounded corners,
            font=\small,
            minimum height = \columnwidth/8,
},
        encoder/.style={
            rectangle,
            draw=black,
            thick,
            fill=myblue,
            align=center,
            rounded corners,
            font=\small,
            minimum height = \columnwidth/8,
},
        encoder_layer/.style={
            rectangle,
            draw=black,
            thick,
            fill=myblue!70,
            align=center,
            rounded corners,
            font=\small,
            minimum height = 0.3cm,
            minimum width=\textwidth/5,
        },
        weighted_avg/.style={
            circle,
            draw=black,
            thick,
            fill=myblue,
            align=center,
            font=\small,
            minimum size=1cm,
        },
        declare function={
            excitation(\t,\w) = sin(\t*\w);
            noise = rnd - 0.5;
            source(\t) = excitation(\t,20) + noise;
            filter(\t) = 1 - abs(sin(mod(\t, 50)));
            speech(\t) = 1 + source(\t)*filter(\t);
        }
    ]
\draw node[llm, minimum width=\textwidth/2.5] (llm0) {LLM (f)};
        \draw node[lora, right=-0.5cm and -1cm of llm0] (LoRA) {$\phantom{LoRA}$\;LoRA};
        \draw node[llm, minimum width=\textwidth/2.5] (llm) {LLM (f)};

\foreach \x in {1,...,5}{
            \draw node[ttoken, above right = 0.3cm and -\x.5cm of llm] (otd\x) {$\phantom{T1}$};
            \draw[myarrow] (llm.north -| otd\x.south) -- (otd\x.south);
        }
        
\draw node[audiobox, above = 0.25cm of otd3] (output) {\small it is so easy to make new friends};

\foreach \x in {0,...,2}{
            \draw node[ttoken, below right = 0.3cm and -\x.99cm of llm] (it\x) {$\phantom{T1}$};
            \draw[myarrow] (it\x.north) -- ( it\x.north |- llm.south);
        }
        
\draw node[llm, minimum width=\textwidth/5, minimum height=0.5cm, below = 0.5cm of it1] (emb) {Text embedding (f)};
        
\foreach \x in {0,...,2}{
            \draw node[atoken, below left = 0.3cm and -\x.99cm of llm] (at\x) {$\phantom{T1}$};
            \draw[myarrow] (at\x.north) -- ( at\x.north |- llm.south);
        }
        
        \foreach \x in {0,...,2}
            \draw[myarrow] (emb.north -| it\x.south) -- (it\x.south);
            
\draw node[cnn, minimum width=\textwidth/6, minimum height=0.5cm, below = 0.5cm of at1] (cnn) {CNN};
        \draw[myarrow] (emb.north -| at1.south) -- (at1.south) node[midway,left] {\small 12.5Hz};
        \draw[myarrow] (emb.north -| at1.south) -- (at1.south) node[midway,right] {\small 80ms};

\draw node[cnn, minimum width=\textwidth/6, minimum height=0.5cm, below = 0.5cm of cnn] (wa) {Weighted average};
        \draw[myarrow] (wa.north -| cnn.south) -- (cnn.south) node [midway, left] {\small 25Hz};
        \draw[myarrow] (wa.north -| cnn.south) -- (cnn.south) node [midway, right] {\small 40ms};
            
\draw node[encoder, minimum width=\textwidth/4.5, minimum height=1.5cm, below = 0.5cm of wa] (encoder) {Audio encoder (f)};
        \draw[myarrow, rounded corners=1.5pt] ($(encoder.west) + (0,0.5cm)$) -- +(-0.75,0) node [near start, above] {$w_L$} -- +(-0.75,0.5) -- +(2,0.5) -- (wa.south);
        \draw[myarrow, rounded corners=1.5pt] (encoder.west) -- +(-0.875,0) node [near start, above] {$\vdots$}  -- +(-0.875,1) -- +(2,1) -- (wa.south);
        \draw[myarrow, rounded corners=1.5pt] ($(encoder.west) + (0,-0.5cm)$) -- +(-1,0) node [near start, below]  {$w_1$} node [near start, above] {$\,\vdots$} -- +(-1,1.5) -- +(2,1.5) -- (wa.south);
        
        \draw node[audiobox, below = 0.5cm of encoder] (speech) {
            \begin{tikzpicture}
                \draw[blue, thick, solid, x=0.009cm, y=0.25cm] (0,1) -- plot [domain=0:360, samples=114, smooth] (\x,{speech(\x)});        
            \end{tikzpicture}
        };
        \draw[myarrow] (speech.north) -- (encoder.south)  node [midway, left] {\small 16kHz};
        
\draw node[textbox, below = 0.5cm of emb] (text-llm) {\small Perform speech\\\small recognition on the\\\small preceding audio\\\small (in English): };
        \draw[myarrow] (text-llm.north) -- (emb.south);

    \end{tikzpicture}\clearpage{}
    \caption{
    \normalsize{Model architecture.
    The audio components are in blue, while the text components are in green.
    We feed a weighted average of the output of all the layers of the pre-trained audio encoder (Eq.~\ref{eq:weighted_average}) to a 1-dimensional CNN to adapt them to the token embedding space. In terms of trainable parameters, we use (f) to represent frozen weights. We do not freeze the (optional) LoRA weights in the LLM if we use them.
    }
    }
    \label{fig:model_architecture}
\end{figure}
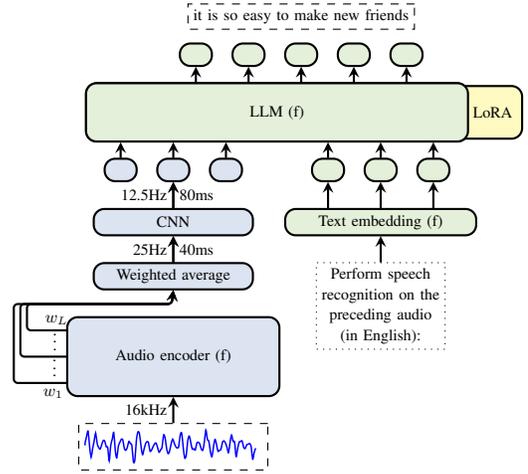

\autoref{fig:model_architecture} shows the model architecture, which consists of a pre-trained speech encoder, an LLM for text generation, and a convolutional network that adapts the audio representations to the text embedding space of the LLM.

\subsection{Speech encoder}
We pre-train a large multilingual speech encoder based on the Conformer \cite{gulati2020conformer} architecture using BEST-RQ \cite{chiu2022self}. We use 16 individual codebooks, with a vocabulary size of 8192 and a dimension of 16. The encoder has 24 layers, each with 8 attention heads, and a feature dimension $d_{AE}$ equal to 1024, for a total of 630M parameters. We use a pre-training mask span of 10 with total effective masking ratio of about 40\%. We use the transformer learning rate scheduler with a peak value of 0.0005 and 50k steps for warm-up, together with AdamW \cite{loshchilov2018decoupled} with weight decay of 0.01. The model is trained for 500k steps in total. The downsampling factor of the model is 4, and the frame length of the output representations is 40ms. The pre-training dataset contains 180K hours of audio from 133 languages, including all the languages that we use for testing in this work.

\subsection{Multilingual LLM}
We use the mT0 family of models \cite{muennighoff2022crosslingual}, an encoder-decoder architecture based on mT5 \cite{raffel2020exploring,xue2020mt5}. The mT0 models have been pre-trained and fine-tuned on over 100 languages
\cite{muennighoff2022crosslingual}.
We use the checkpoints available in Huggingface \cite{wolf2019huggingface}  (mT0-XL\footnote{\texttt{\url{https://hf.co/bigscience/mt0-xl}}}: 3.7B parameters, hidden dimension $d_{LLM}=2048$ and mT0-XXL\footnote{\texttt{\url{https://hf.co/bigscience/mt0-xxl}}}: 12.9B parameters, hidden dimension $d_{LLM}=4096$, respectively).

\subsection{Adapter module} 
We use a 2-layer 1-dimensional CNN between the speech encoder and the LLM \cite{das2024speechverse}. We use a kernel size $k=3$ on both layers. The first layer performs the upsampling in the feature dimension to $d_{LLM}$ and downsampling in the time dimension (40ms to 80ms in frame lengths). 
The second layer adds depth to the adapter. 
The module has 18.9M and 62.9M parameters (initialized randomly) when used with mT0-XL and mT0-XXL, respectively.

\subsection{Full model}
To connect the speech encoder representations to the LLM, we use a weighted average of the output $\bm{h}_{AE}^l$ of all the layers in the speech encoder, such that:
\begin{equation}
    AE(\bm{x}) = \frac{1}{L}\sum_{l=1}^L w_l\bm{h}_{AE}^l(\bm{x}_{l-1}) \;\in\;\mathbb{R}^{d_{AE}\times t},\label{eq:weighted_average}
\end{equation}
where $\bm{x}$ is the audio, $\bm{x}_l = \bm{h}_{AE}^l(\bm{x}_{l-1})$ is the output of layer $l$, $\{w_l, \, l\in{1,\ldots,L}\}$ is a set of learnable parameters, and $t$ is the length of the sequence.
$AE(\bm{x})$ is then passed to the CNN module, such that $A(\bm{x}) = CNN(AE(\bm{x}))\in\mathbb{R}^{d_{LLM}\times t/2}$. We concatenate these audio sequences with the tokenized text that has been passed through the trained token embedding layer of the LLM, therefore soft-prompting the LLM with audio representations.

\subsection{Multi-task training}
\noindent\textbf{Tasks} To evaluate the cross-lingual performance of our models for zero-resource ST and ASR, we train our models using either ST and ASR data. When training the model with both ST and ASR data, we are implicitly requesting the model to learn how to perform both tasks at the same time. To achieve this, we use next-token prediction with cross-entropy loss as implemented by Huggingface for both tasks and change the prompt that we use to instruct the model depending on the task.

\noindent\textbf{Train prompts} The tokenized text represents an instruction for the model that is task-dependent. In preliminary experiments, we observed better robustness and improved performance when using several prompts at training time. Therefore, for ST we use 25 paraphrased prompts similar to the following instruction: ``Perform speech translation into English using the preceding audio: ''.
For ASR, we use 25 paraphrased prompts similar to the following instruction: ``Perform speech recognition in \{language\} using the preceding audio: '', where \{language\} is the language of the input speech. These prompts are sampled uniformly at random conditional on the task.

\noindent\textbf{Test prompts} We use the following prompts at test time: for ST, we use ``Transcribe the content of this audio into English in textual form: '', and for ASR, we use ``The preceding audio is in \{language\}. Perform speech recognition (in \{language\}): ''.

\subsection{Adapter Training}
To align speech and text representations, we employ a CNN adapter for speech and incorporate LoRA~\cite{hu2021lora} for text LLMs, as outlined in \cite{gong2023listen}. These components are jointly trained on speech translation and ASR data. To further improve the alignment, prior to the joint training of adapters, we pre-train the CNN adapter on ASR data. This pre-training in turn facilitates a better initialization for joint training. In the following section, we assess the impact of training these adapters in a multi-stage manner. In LoRA, we use $\alpha=10$ and rank $r=16$, resulting in 9.4M and 18.8M additional parameters for mT0-XL and mT0-XXL models.

\section{Zero-resource Speech Translation} \label{sec:experiments}
\begin{table*}[t]
    \centering
    \caption{\normalsize{Speech translation results. The numbers show BLEU scores in the CoVoST2 dataset. The model has not seen paired audio/text data for Dutch (nl) nor Catalan (ca). \textbf{Bold face} shows the best performance in zero-resource for the two different model sizes.}}
    \vspace{-0.4cm}
    \clearpage{}\setlength{\tabcolsep}{2pt}
\footnotesize
\begin{tabular}{ccl ccc cccccc c ccc}
\toprule
& & & & & & \multicolumn{6}{c}{\textbf{Trained language pairs}} & & \multicolumn{3}{c}{\textbf{Zero-resource pairs}} \\
\cmidrule{7-12}\cmidrule{14-16}
\textbf{Row} & \textbf{Experiment} & \textbf{Model} & \makecell{\textbf{Pretraining}} & \makecell{\textbf{Trainable} \\ \textbf{Params}} & \textbf{Tasks} & \makecell{\textbf{fr→en}\\ (14,760)} & \makecell{\textbf{de→en}\\ (13,511)} & \makecell{\textbf{es→en}\\ (13,221)} & \makecell{\textbf{it→en}\\ (8,951)} & \makecell{\textbf{pt→en}\\ (4,023)} & \makecell{\textbf{Mean}\\ (PFIGS)} & & \makecell{\textbf{nl→en}\\ (1,699)} & \makecell{\textbf{ca→en}\\ (12,730)} &  \makecell{\textbf{Mean}} \\
\midrule

1 & \multirow{3}{*}{\makecell{Text-only\\ LLMs}} & mT0-XL & \multirow{3}{*}{None} &   -   &  -   & 31.70 & 29.53 & 32.76 & 29.89 & 40.55 & 32.89 & & 35.61 & 24.81 & 30.21 \\
2 & & mT0-XXL &              &  - & -  & 35.70 & 33.80 & 36.73 & 34.10 & 46.50 & 37.37 & & 39.44 & 29.09 & 34.27 \\
 3 & & mT0-XL-FT   &  & \textit{LoRA} & MT & 46.90 & 40.48 & 46.62 & 43.15 & 54.00 & 46.23 & & 43.79 & 36.78 & 40.28 \\
\midrule
4 & \multirow{6}{*}{\makecell{SLMs \\ (Ours)}} & mT0-XL    & \xmark & \textit{CNN only} & ST \& ASR & 29.46 & 23.07 & 29.40 & 25.49 & 30.67 & 27.62 & & 18.01 & 20.42 & 19.22 \\ 5 & & mT0-XL & \xmark & \textit{CNN + LoRA} & ST \& ASR & 31.81 & 25.91 & 32.05 & 27.82 & 30.88 & 29.69 & & 16.02 & 21.75 & 18.88 \\

6 & & mT0-XL   & \cmark & \textit{CNN + LoRA} & ST \& ASR & 33.05 & 27.41 & 33.27 & 28.62 & 33.60 & 31.19 &     &     \textbf{18.23} &          22.89 & 20.56 \\ \cmidrule{3-16}
7 &  & mT0-XL   & \cmark & \textit{CNN + LoRA} & ST     & 33.78 & 28.10 & 34.10 & 29.60 & 34.66 & 32.05 & & 17.86 & 23.50 & \textbf{20.68} \\ 8 & & mT0-XL-FT & \cmark & \textit{CNN + LoRA} & ST & 34.42 & 29.14 & 34.61 & 30.58 & 35.75 & 32.90 & & 17.49 & \textbf{23.81} & 20.65 \\ 

\cmidrule{3-16}
9 &  & mT0-XXL  & \cmark & \textit{CNN + LoRA} & ST \& ASR & 35.57 & 32.41 & 37.08 & 34.23 & 45.17 & 36.89 & & \textbf{23.26} & \textbf{25.78} & \textbf{24.52} \\ \bottomrule
\end{tabular}
\clearpage{}
    \label{tab:st_results}
\end{table*}

\subsection{Datasets}
For Speech Translation (ST), we used both ASR and ST data from CoVoST2 \cite{wang2020covost} and Europarl-ST \cite{iranzo2020europarl}. The training data was limited to the following languages: Portuguese (pt), French (fr), Italian (it), German (de), Spanish (es), and English (en). The X$\rightarrow$en (X $\in$ PFIGS) ST training data comprised approximately 800 hours of data. Additionally, the ASR partition of the same data added another 800 hours, totaling around 1600 hours of training data. The evaluations were conducted solely on the CoVoST2 dataset, focusing on the Catalan-English (ca$\rightarrow$en) and Dutch-English (nl$\rightarrow$en) language pairs. These language pairs are strategically selected for zero-resource testing due to their linguistic characteristics. Catalan, being a Romance language similar to French, Italian, and Spanish, and Dutch, sharing similarities with German, allowed us to effectively assess cross-lingual transfer capabilities of the models. We report BLEU scores~\cite{post-2018-call} to evaluate speech translation quality.

\subsection{Experiments}
\begin{figure*}
    \includegraphics[width=\textwidth]{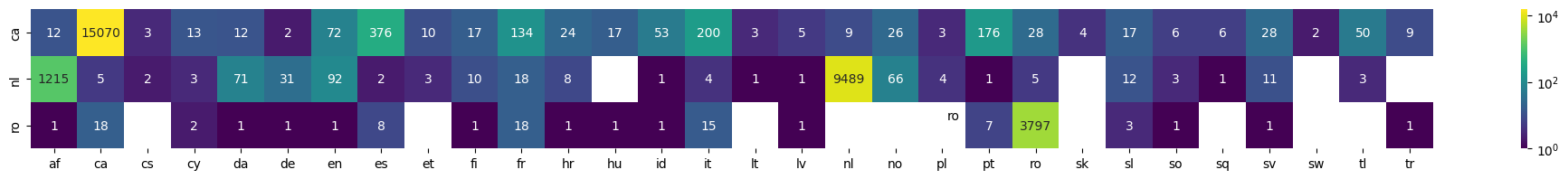}
\caption{\normalsize{Language confusions for the ASR test languages in Common Voice.}}
    \label{fig:language_confusions}
\end{figure*}

Prior to assessing the performance of SLMs for speech translation on zero-resource language pairs, we establish an upper bound (i.e., text-only translation) for this work by utilizing the same LLMs employed for training the SLMs. Two different models are evaluated: mT0-XL and mT0-XXL. Furthermore, we include a fine-tuned version of mT0-XL, denoted as mT0-XL-FT, which is trained (with LoRA) on the same text translation data as used for training the SLMs.

We trained several variants of SLMs to evaluate the effectiveness of the parameterization techniques, and pre-training strategies for zero-resource cross-lingual transfer learning in ST. 

The work conducted by authors in \cite{wang2023slm} demonstrated that a 2-layer transformer adapter with random subsampling achieves comparable BLEU scores to previous state-of-the-art work in ST. Inspired by this approach, we use light-weight adapters for training SLMs but disentangle the adaptation from representation learning. We utilize CNN parameters for both subsampling and encoding speech representations into textual space, while LoRA parameters are employed for joint learning and decoding. Additionally, we perform pre-training of the CNN layers before fine-tuning both CNN and LoRA parameters together. This pre-training stage aims to enhance the encoding of speech features into textual space and stabilize the training process before introducing LoRA parameters.

\subsection{Results}

Table \ref{tab:st_results} presents the speech translation results from both text-only LLMs and our spoken language models (SLMs). For the text-only LLMs, we utilized the original ground-truth transcriptions as input. Firstly, comparing rows 1 and 2, the performance of mT0-XXL significantly outperforms mT0-XL across all language pairs, likely due to its increased model size and capacity. Additionally, we observe that mT0-XL-FT, which underwent additional fine-tuning, achieved a 14\% absolute improvement across trained language pairs and a 10\% improvement across zero-resource pairs. The performance of these models is presented solely to serve as a topline reference that can be achieved with these LLMs. However, it is important to note that these public text-only models may have already encountered the data for zero-resource language pairs, and their naming conventions do not carry significant implications. 

\noindent\textbf{Impact of CNN subsampling with LoRA adaptation} We compare the performance of various spoken language models trained following the methodology we presented in Section \ref{sec:methodology}. Rows 4 and 5  of \autoref{tab:st_results} show the results of trained language paris with and without LoRa adaptation. We observe a notable improvement with the introduction of LoRA in addition to CNN subsampling. However, the performance of zero-source pairs degraded slightly, likely due to LoRA parameters being fine-tuned before the CNN features could properly encode speech features, resulting in suboptimal performance for unseen languages. Comparing rows 5 and 6, where we performed additional pre-training of CNN layers before fine-tuning CNN and LoRA, the performance of both trained language pairs and zero-resource language pairs improved significantly. This clearly demonstrates that while LoRA adaptation benefits spoken language models, sequential training with pre-training of CNN layers is crucial for optimal performance.

\noindent\textbf{Multi-task training} By comparing the results from rows 6 and 7, we evaluate the performance of speech translation as a standalone task by omitting the ASR task. In contrast to the findings reported in  \cite{rubenstein2023audiopalm}, our observations indicate that incorporating ASR data to train the model in a multi-task fashion does not improve performance across all language pairs. Instead, we notice a slightly lower average performance for both seen and unseen languages during training when ASR data is included. This result suggests that if the speech translation (ST) data is sufficient to train the adapter to map the audio representations to the token embedding space, additional ASR data may not be necessary. The results in row 8 indicate that further fine-tuning of the text-only LLM did not lead to significant performance improvements in the spoken language model (SLM) task.

\noindent\textbf{Zero-resource performance}
As expected, we observe lower zero-resource performance for the models compared to seen languages. However, the larger-scale experiment using mT0-XXL demonstrates that speech translation quality can be comparable to text-based translation in certain scenarios: for Catalan, the ST BLEU score with mT0-XXL (row 9) is remarkably only 4\% absolute lower than the text-only translation BLEU score (row 2). This result, however, does not translate to Dutch, where we observe a greater degradation of 16\%. We hypothesize that two factors contribute to the worse results for Dutch: (1) Dutch has less phonetic overlap with the languages in the training set (max Jaccard similarity for phonemes in PFIGS+en languages and Dutch is 0.48 for German, while Catalan's is 0.79 with Spanish), making generalization more challenging, and (2) as observed in multilingual models, different languages require different training times to be fitted adequately (e.g., in \cite{bai2024efficient}, the authors discuss how Dutch is under-fitted while Portuguese, a Romance language, is already fitted).

\section{Zero-resource ASR}

\subsection{Datasets}
For the ASR task, we utilized data from various sources, including Common Voice v14.0 \cite{ardila2019common}, VoxPopuli \cite{wang2021voxpopuli} (labeled portion only), FLEURS \cite{conneau2023fleurs}, LibriVox-MLS \cite{pratap2020mls}, and the ASR data from Europarl-ST (since Common Voice and CoVoST2 have significant overlap). For the experiments described in Section \ref{ssec:asr_experiments}, we created several variants of the ASR datasets, each containing different sets of languages seen during training. However, we use Catalan (ca), Dutch (nl), and Romanian (ro) as the test languages for evaluation. For metrics, we use word error rate (WER). We perform basic normalization of the references (removing punctuation, lowercase, etc.), and directly compare to the output of the LLM.

\subsection{Experiments}
\label{ssec:asr_experiments}
We begin by presenting a topline system as a reference, which includes all zero-resource languages during training. Subsequently, we describe a set of experiments to investigate the model's behavior in the zero-resource scenario under the following conditions: (1) using sequential training, (2) increasing the amount of training data, and (3) varying the number of languages present in the training sets. For all the ASR experiments, we utilize the mT0-XXL model.

\noindent\textbf{ASR topline}
Similar to the setup in ST, we use PFIGS+en languages for training, sampling up to 500hrs per language from Common Voice, VoxPopuli, FLEURS, and including CoVoST2, Europarl-ST, and Europarl-ST's ASR data. For the topline, we add up to 350hrs of Catalan, Dutch, and Romanian data, totaling 4,400hrs of speech data.

\noindent\textbf{ASR experiment 1 (ASR-E1)}
To study how the model performs in the zero-resource setting, we remove Catalan, Dutch, and Romanian data from the topline dataset, totaling 3,500hrs of audio. We train models with and without LoRA adaptation.

\noindent\textbf{ASR experiment 2 (ASR-E2)}
In preliminary results, we observe that the model does not always output the correct language in the ASR task. Therefore, to understand the role of the dataset size and number of training languages, we create a dataset with 36 languages from Common Voice, Voxpopuli, FLEURS, and LibriVox-MLS. We cap the number of hours for each language between 350hrs and 20hrs, for a total of 50 languages. Out of these 50 languages, we keep 36 that were used in the training of mT0 (not including Catalan, Dutch, and Romanian). We also add the ST data from CoVoST2 and ST+ASR data from Europarl-ST, for a total of about 9,000hrs of data.

\noindent\textbf{ASR experiment 3 (ASR-E3)}
To further study the impact of the number of languages in performance and language confusion, we follow the same steps as in ASR-E2, but create a dataset with 84 languages by capping the maximum and minimum number of hours to 350hrs and 10hrs, respectively. We obtain 67 languages that overlap with mT0 training. After adding ST data, we obtain about 9,000hrs of training data.

\subsection{Results}

Table \ref{tab:zero-resource_asr} presents the results on zero-resource speech recognition across different experiments. 

\noindent\textbf{Language confusion}
We initially observed that the model does not consistently output text in the desired language. Figure \ref{fig:language_confusions} presents a confusion matrix for all three zero-resource languages against all other languages used in training. While Catalan is confused with Spanish, Italian, and Portuguese, Dutch is highly confused with Afrikaans. To quantify this behavior, we employ \texttt{langdetect} \cite{langdetect} to detect the output language and report the language accuracy for each test language and experiment in \autoref{tab:zero-resource_asr}. Even in the topline model, where we included the test languages during training, the language accuracy ranges from 89\% (Dutch) to 98\% (Romanian). In the zero-resource scenarios, the language accuracy is further reduced. To decouple the effect of language confusions from the ASR capabilities, we filter out the hypotheses in the wrong language and re-score, presenting both the unfiltered and filtered results.

\noindent\textbf{Sequential training} 
In ST task, we observed that sequential training with pre-training of CNN layers is crucial for optimal performance. However, for the experimental results presented in ASR-E1, we observe that using LoRA adaptation in addition to CNN subsampling degrades the performance of the model. This can be explained by the model losing generalization capabilities to unseen languages, giving the worst results across our experiments.

\noindent\textbf{Larger scale} We finally show the results for the experiments where we use more hours of data and more languages in the training set (36 languages in ASR-E2 and 67 languages in ASR-E3). In both of these experiments we observe WER improvements over the previous experiments. In ASR-E3, we observe that the language confusion problem is reduced, reaching the same language accuracy for both Catalan and Romanian than the baseline. However, this is not reflected in the WER, where the model trained in ASR-E2 shows the best WER performance.

\begin{table}
    \centering
    \caption{\normalsize{Zero-resource ASR results in Common Voice. The language accuracy shows the percentage of hypotheses generated in the prompted language. $^\dagger$ represents the WER calculated by first filtering out the hypotheses that are in the wrong language.}}
    \vspace{-0.4cm}
    \clearpage{}\setlength{\tabcolsep}{2pt}
\footnotesize
\begin{tabular}{ccl|cccc|ccccc}
    \toprule
    \multirow{2}{*}{\textbf{Experiment}} & \multirow{2}{*}{\textbf{LoRA}} & \multirow{2}{*}{\textbf{Eval.}} & \multicolumn{4}{c}{\makecell{\textbf{Language} \textbf{Accuracy}}} & \multicolumn{4}{c}{\textbf{WER}} \\
     &  &  & \textbf{ca} & \textbf{nl} & \textbf{ro} & \textbf{Mean} & \textbf{ca} & \textbf{nl} & \textbf{ro} & \textbf{Mean} \\
\midrule
     \multirow{1}{*}{Topline} & \xmark & All            & 0.92 & 0.89 & 0.98 & 0.93 & 25.6 & 13.9 & 26.9 & 22.1 \\
     \midrule
\multirow{5}{*}{ASR-E1} & \xmark & All              & 0.91 & 0.80 & 0.89 & 0.87 & 44.7 & 37.8 & 64.2 & 48.9 \\
                             & \xmark & Subset$^\dagger$ &  --  &  --  &  --  &  --  & 41.3 & 31.6 & 60.4 & 44.4 \\
                             \cmidrule{2-11}
                             & \cmark & All              & 0.57 & 0.28 & 0.73 & 0.53 & 51.8 & 75.5 & 57.9 & 61.8\\
                             & \cmark & Subset$^\dagger$ &  --  &  --  &  --  &  --  & 31.6 & 39.0 & 45.8 & 38.8\\
    \midrule
     \multirow{2}{*}{ASR-E2} & \xmark & All              & 0.91 & 0.83 & 0.97 & 0.90 & 48.7 & 32.3 & 52.5 & 42.4 \\
                             & \xmark & Subset$^\dagger$ &  --  &  --  &  --  &  --  & 41.4 & 29.6 & 37.1 & 36.0 \\
    \midrule
     \multirow{2}{*}{ASR-E3} & \xmark & All              & 0.92 & 0.86 & 0.98 & 0.92 & 53.6 & 30.5 & 50.1 & 42.6 \\
                             & \xmark & Subset$^\dagger$ &  --  &  --  &  --  &  --  & 38.1 & 28.2 & 43.6 & 36.6 \\
    \bottomrule
\end{tabular}

\clearpage{}
    \label{tab:zero-resource_asr}
\end{table}

\section{Conclusions}
In this paper, we propose a method to perform zero-resource speech translation and automatic speech recognition. We use a pre-trained speech encoder and a multilingual LLM, and show that training a lightweight adapter between the output of the speech encoder and the token embedding space of the LLM allows to perform speech tasks in languages for which the model has never seen paired audio-text data.

\bibliographystyle{IEEEtran}
\bibliography{references}

\end{document}